\documentclass[letterpaper, 10 pt, conference]{ieeeconf}  
\IEEEoverridecommandlockouts
\pdfoutput=1

\usepackage{cite}
\usepackage{amsmath,amssymb,amsfonts}
\usepackage{algorithmic}
\usepackage{graphicx}
\usepackage{textcomp}
\usepackage{xcolor}
\usepackage{todonotes}
\usepackage{hyperref}
\usepackage{graphics} 
\usepackage{epsfig}
\usepackage{times} 
\usepackage{bm}
\usepackage{multirow}
\usepackage{arydshln}
\usepackage{booktabs}

\usepackage{float}

\hypersetup{
    colorlinks=true,
    linkcolor=blue,
    filecolor=magenta,      
    urlcolor=magenta,
    pdftitle={Overleaf Example},
    pdfpagemode=FullScreen,
    }

\usepackage[protrusion=true,
             expansion=true,
             tracking=true,
             final,
             stretch=25,factor=1250]{microtype}

\def\BibTeX{{\rm B\kern-.05em{\sc i\kern-.025em b}\kern-.08em
    T\kern-.1667em\lower.7ex\hbox{E}\kern-.125emX}}

\usepackage[capitalize]{cleveref}
\crefname{section}{Sec.}{Secs.}            
\Crefname{section}{Section}{Sections}
\Crefname{table}{Table}{Tables}
\crefname{table}{Tab.}{Tabs.}

\usepackage{color}
\definecolor{darkgreen}{rgb}{0, 0.7, 0}
\newcommand{\greenbf}[1]{\textcolor{darkgreen}{\bf #1}}
                                   
\Large
\begin{document}
\title{\Large \bf
FusionRCNN: LiDAR-Camera Fusion for Two-stage 3D Object Detection
}

\author{{Xinli Xu$^{1}$, Shaocong Dong$^{1}$, Lihe Ding$^{1}$, Jie Wang$^{1}$, Tingfa Xu$^{1}$, Jianan Li$^{1}$}
\thanks{$^{1}$Beijing Institute of Technology, Beijing, CN, \{xxlbigbrother, dscdyc1010295799, dean.dinglihe, jwang123bit\}@gmail.com, \{ciom xtf1, lijianan\}@bit.edu.cn}
}
\maketitle

\begin{abstract}

3D object detection with multi-sensors is essential for an accurate and reliable perception system of autonomous driving and robotics. Existing 3D detectors significantly improve the accuracy by adopting a two-stage paradigm which merely relies on LiDAR point clouds for 3D proposal refinement. Though impressive, the sparsity of point clouds, especially for the points far away, making it difficult for the LiDAR-only refinement module to accurately recognize and locate objects.To address this problem, we propose a novel multi-modality two-stage approach named FusionRCNN, which effectively and efficiently fuses point clouds and camera images in the Regions of Interest (RoI). 
FusionRCNN adaptively integrates both sparse geometry information from LiDAR and dense texture information from camera in a unified attention mechanism.
Specifically, it first utilizes RoIPooling to obtain an image set with a unified size and gets the point set by sampling raw points within proposals in the RoI extraction step; then leverages an intra-modality self-attention to enhance the domain-specific features, following by a well-designed cross-attention to fuse the information from two modalities.FusionRCNN is fundamentally plug-and-play and supports different one-stage methods with almost no architectural changes. Extensive experiments on KITTI and Waymo benchmarks demonstrate that our method significantly boosts the performances of popular detectors.Remarkably, FusionRCNN significantly improves the strong SECOND baseline by 6.14\% mAP on Waymo, and outperforms competing two-stage approaches. Code will be released soon at \url{https://github.com/xxlbigbrother/Fusion-RCNN}.

\end{abstract}

\section{Introduction}
3D object detection is one of the fundamental tasks in autonomous driving and robotics, which aims to capture accurate 3D information with multiple sensors. Since LiDAR sensors enjoy the natural advantage of obtaining accurate depth and shape information, previous methods achieve competitive performance by  using only point clouds. Furthermore, some attempts significantly improve the performance through a two-stage refinement module, which inspires the researchers to explore more effective LiDAR-based two-stage detectors. 

\begin{figure}

    \centering
    \includegraphics[width=8.5cm]{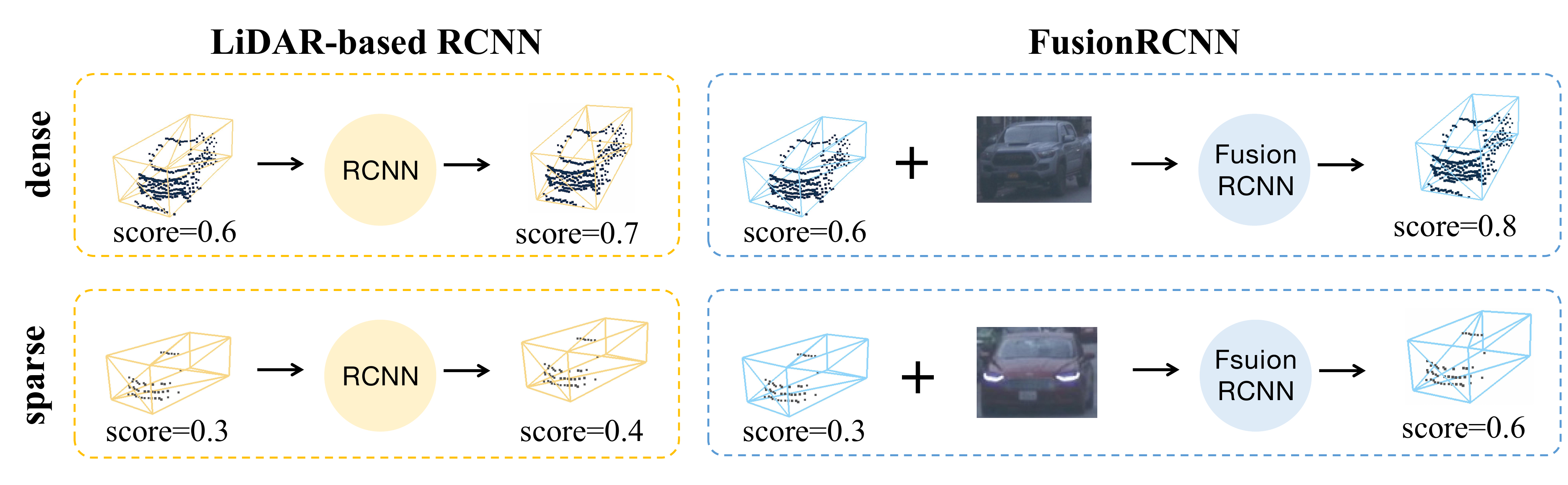}
    \caption{Comparison of our method with previous LiDAR-based two-stage methods. When objects comprise sparse point clouds, LiDAR-based methods fail to correctly determine the category and give less confident scores, while our method effectively combines point cloud structure with dense image information to solve such problems.}
    \label{fig:intro}
\vspace{-0.4cm}
\end{figure}

Two-stage methods can be divided into three main categories based on the representation of Point of Interest, \emph{i.e.}, point-based, voxel-based and point-voxel-based. 
Point-based approaches~\cite{shi2019pointrcnn,yang2019std, li2021lidar, sheng2021improving} take the sampling points as input, and obtain point features for RoI refinement.
Voxel-based methods~\cite{shi2020points,deng2021voxel} rasterize point clouds into voxel-grids and extract features from 3D CNNs for refinement.
Point-Voxel-based approaches\cite{shi2020pv,shi2021pv} combine the two types of feature learning schemes to improve detection performance. However, no matter for what representation, the sparsity and non-uniform distribution characteristics of point clouds make it difficult to distinguish and locate objects in the far distance, leading to false or missed detections, as illustrated in Fig.~\ref{fig:intro}. Things get extremely worse when the proposals contain few (1-5) points, from which we can hardly obtain enough semantic information. Fortunately, camera is complementary to LiDAR by providing dense texture information. How to design the LiDAR-Camera fusion paradigm in two-stage to well leverage their complementary strengths is of great importance.

In this work, we focus on fusing LiDAR point clouds and images in the refinement stage. 
Previous works~\cite{xie2020pi} utilize an image segmentation sub-network to extract image features and attach image features to the raw points.
However, we find that the point-based fusion ignores semantic density of image features and heavily relies on the image segmentation sub-networks.
In light of the above, this work presents a deep fusion method, dubbed FusionRCNN, which comprises three steps:
i) extract RoI features from points and images corresponding to proposals from any one-stage detectors; ii) fuse the features of these two modalities through well-designed intra-modality self-attention and inter-modality cross-attention, abandoning the heavy reliance on hard-associations between points and images while keeping the semantic density of images; iii) feed the encoded fusion features into a transformer-based decoder to predict the refined 3D bounding boxes and confidence scores.

Our FusionRCNN is generic and can significantly boosts the detection performance. 
Extensive experiments on KITTI~\cite{geiger2013vision} and Waymo~\cite{sun2020scalability} demonstrate that our FusionRCNN brings obvious performance gain upon LiDAR-only methods, especially for difficult samples with sparse point clouds (Hard level on KITTI and $ 50m\mathrm{-Inf} $ on Waymo). Remarkably, applying our two-stage refinement network on SECOND~\cite{yan2018second} baseline improves the detection performance by 11.88 mAP in the range of $ \ge 50m $ (46.93 $\rightarrow$ 58.81 mAP on Vehicle) on Waymo.

\begin{figure*}
\begin{center}
\includegraphics[width=16cm]{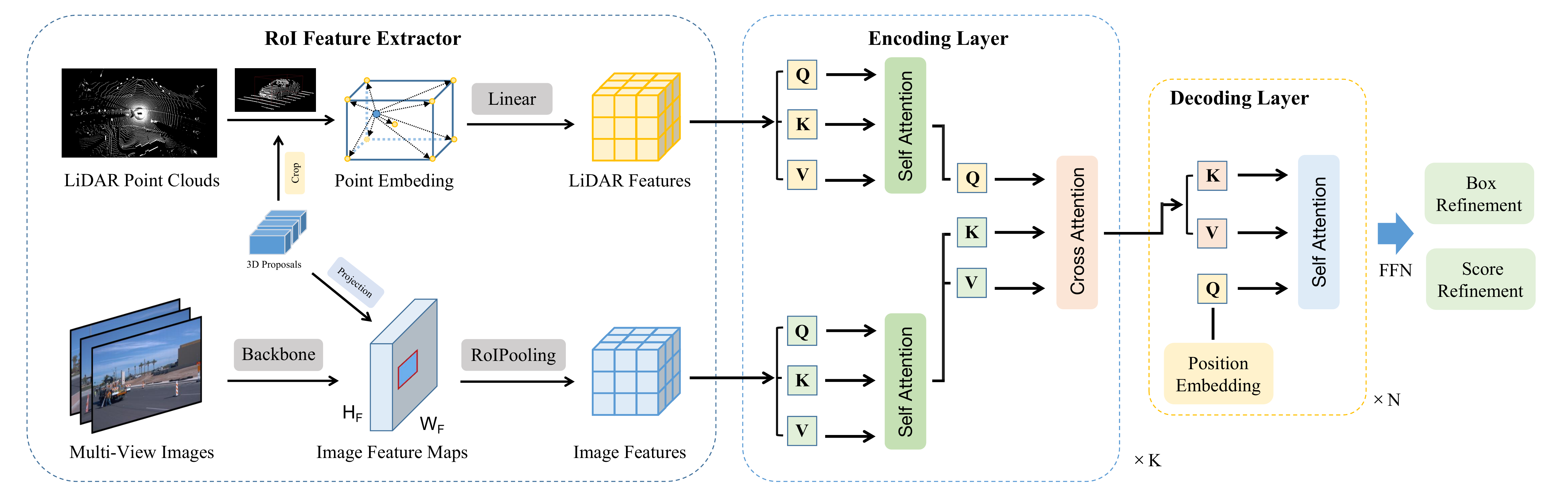}
\end{center}
   \caption{Overall architecture of FusionRCNN. Given 3D proposals, LiDAR and image features are extracted separately through RoI feature extractor. Then, the features are fed into $K$ fusion encoding layers which comprises self-attention and cross-attention modules. Finally, point features fused with image information are further fed into a decoder and predict the refined 3D bounding boxes and confidence scores.}
\label{fig:framework}
\end{figure*}

To sum up, this work makes the following contributions:
\begin{itemize}
    \item We propose a flexible and effective two-stage multi-modality 3D detector named FusionRCNN, which fuses image and point clouds in regions of interest and can boost existing one-stage detectors with minor changes.
    \item We utilize a novel transformer-based mechanism to simultaneously achieve attentive fusion between pixel set and point set, which is robust to calibration noise.

    \item Our method has superior performance compared to two-stage approaches on KITTI and Waymo Open Dataset, especially on difficult samples with sparse points.
\end{itemize}

\section{Related Works}
\noindent\textbf{LiDAR-Based 3D Detection:} 
Existing LiDAR-based 3d detection methods can be broadly grouped into three categories, The Voxel-based, Point-based, and Range View. Voxel-based detetors voxelize the unstructured point clouds as a regular 2D/3D grid which conventional CNNs can be easily applied. The pioneer work MV3D~\cite{chen2017mv3d} projects the point clouds to 2D bird-eye view grids and places lots of predefined 3D anchors for generating high accurate 3D candidate boxes, motivating following efficient bird-eye view representation methods. VoxelNet~\cite{zhou2018voxelnet} applies mini PointNet~\cite{qi2017pointnet} for voxel featurea extraction. SECOND~\cite{yan2018second} introduces 3D sparse convolution to accelerate 3D voxel processing. For Point-based methods, PointNet and its variants~\cite{qi2017pointnet++} directly take the raw points as input and use symmetric operators to address the unorderness of point clouds. PointRCNN~\cite{shi2019pointrcnn} and STD~\cite{yang2019std} segment foreground points with PointNet and generate proposals. 3DSSD~\cite{yang20203dssd} proposes a new sampling strategy for efficient computation. Range View detectors~\cite{fan2021rangedet,sun2021rsn} represent LiDAR point clouds as dense range images, where pixels contains extra accurate depth information. Compared to other methods, Voxel-based detetors balances the efficiency and performance, we choose the voxel-based detector as RPN networks in this paper.

\noindent\textbf{LiDAR-Camera 3D Detection:}
Recently, LiDAR-Camera 3D detection has been receiving increasing attention as the two types sensors are complementary. LiDARs provide sparse point clouds containing accurate depth information, while cameras provide high-resolution images containing rich color and textures. MV3D~\cite{chen2017mv3d} creates 3D object proposal from LiDAR bev features and projects the proposals to multi-view images to extract RoI features. F-PointNet~\cite{qi2018frustum} lift images proposal into a 3D frustum and achieve high performance. Point-level fusion methods decorate raw foreground LiDAR points and apply a common LiDAR-based detectors on the decorated point clouds. Among these methods, PointPainting~\cite{vora2020pointpainting}, PointAugmenting~\cite{wang2021pointaugmenting}, MVP~\cite{meyer2019sensor}, FusionPainting~\cite{xu2021fusionpainting} and AutoAlign which have gained great success are input-level decoration, while DeepFusion~\cite{li2022deepfusion} and Deep Continues Fusion are feature-level decoration. Recent works TransFusion~\cite{bai2022transfusion} and FUTR3D~\cite{chen2022futr3d} initialize object queries in 3D space and fuse image feature on the proposals. To our knowledge, few works focus on two-stage fusion networks, in this paper we propose a novel framework which can be applied as a plug-and-play RCNN~\cite{girshick2015fast, ren2015faster} module to existing detectors and boost their performance significantly.

\section{Method}

Given $ M $ predicted proposals containing 3D bounding boxes $\bm{B}= \{\bm{b}_{i}\}_{i=1}^{M} $, where $ \bm{b}_{i} = \{x,y,z,l,h,w,\theta\} $ (box center position, size, and heading angle), and confidence scores $\mathbf{S} = \{s_{i}\}_{i=1}^{M} $ from any one-stage detectors. We aim to improve the detection results based on point clouds $\bm{P}$ and camera images $\bm{I}=\{ I_i \in \mathbb{R}^{3 \times H_I \times W_I}\}_{i=1}^{T}$ from $ T $ views, \emph{i.e.}
\begin{equation}
    (\bm{B}_{r}, \bm{S}_{r}) = \mathcal{R}(\bm{B}, \bm{P}, \bm{I}),
\end{equation}
where $\bm{B}_{r}$ and $\bm{S}_{r}$ are corrected bounding boxes and confidence scores, and $ \mathcal{R} $ represents the proposed network.

\cref{fig:framework} shows the overall architecture of the proposed FusionRCNN. We adopt the RoI Feature Extractor (\cref{RoI extraction module}) to extract the RoI features from points and images corresponding to $\bm{B}$, then fuses the features of these two modalities through Fusion Encoder (\cref{fusion encoding layer}). The encoding fusion features are further fed into Decoder (\cref{decoding layer}) and predict the refined 3D bounding boxes and confidence scores.

\subsection{RoI Feature Extractor}  \label{RoI extraction module}
Start with giving 3D bounding boxes $\bm{B}$, point clouds $\bm{P}$ and camera images $\bm{I}$, in order to capture sufficient structure and context information, we fix the center of the bounding box $ \bm{b}_{i} $ while expanding the length, width and height with radio $ k $, and feed the scaled RoI to the feature extractor. We adopt a two-branch architecture, where the point/image RoI features are extracted from point clouds $\bm{P}$ and images $\bm{I}$ individually.

For the point branch, points within the corresponding box $ \bm{b}_{i} $ after expansion are sampled or padded to a unified number $ N $. Inspired by the point embedding methods used in~\cite{sheng2021improving}\cite{li2021lidar}, We enhance the point features by concatenating the distance to the eight corners and the center of $\bm{b}_{i}$ as

\begin{equation}
    \bm{F}^{P}_{i} = \mathcal{L}({\Delta \bm{p}_{1}, \Delta \bm{p}_{2}, ..., \Delta \bm{p}_{8}, \bm{p}_{b}, \bm{p}_{e}}),
\end{equation}
where $ \Delta \bm{p}_{j} $ is the distance to the $ j $-th corners of the box $ \bm{b}_{i} $ , $ \bm{p}_{b} $ is the center coordinates of the bounding box, $ \bm{p}_{e} $ is extra LiDAR point information like reflectivity, and $ \mathcal{L}(\cdot)$ is a linear projection layer to map point features into an embedding with $ C $ channels. Formally, the point RoI features are $\bm{F}^{P}=\{ \bm{F}^{P}_{i} \in \mathbb{R}^{C \times N}\}_{i=1}^{M}$.

For the image branch, the original multi-view images are converted into feature maps via ResNet~\cite{he2016resnet} and FPN~\cite{lin2017fpn}. We project the expanded 3D bounding boxes onto the 2D feature map, and crop the 2D feature to obtain the image embedding corresponding to the RoI. Specifically, eight 3D corners are projected onto the 2D feature map by the intrinsics and extrinsics of the cameras, from which we calculate the minimum circumscribed rectangle and perform RoI pooling to get the image feature $ \bm{F}_{i}^{I} $ with a unified size $ S \times S $ corresponding to $ \bm{b}_{i} $. Another linear layer finally projects $ \bm{F}_{i}^{I} $ into the same dimension $ C $ as the point features. Formally, the image RoI features are $\bm{F}^{I}=\{ \bm{F}^{I}_{i} \in \mathbb{R}^{C \times S \times S}\}_{i=1}^{M}$.

\subsection{Fusion Encoder}\label{fusion encoding layer}
Based on the above RoI Feature Extractor,
 we can get the per-point feature and the per-pixel image feature (pixel size varies since we fix a $ S \times S $ pooling size while the projected proposal sizes are different)  inside the RoI. Instead of fusing features by painting the image features into points like previous methods~\cite{vora2020pointpainting,wang2021pointaugmenting}, which prefer to utilize the direct correspondence between points and image pixels but neglects the fact that a local region of pixels can contribute to one point and vice versa, we leverage self-attention and cross-attention to achieve the Set-to-Set fusion.
Specifically, to make point and image features align with each other and better model the inner relationships, 
we first feed them into the multi-head self-attention layer respectively.

For embedded point features $ \bm{F}^{P} $, we have
\begin{equation}
    \bm{Q}_{P},\bm{K}_{P},\bm{V}_{P} = \bm{W}_{P}^{Q}\bm{F}^{P},\bm{W}_{P}^{K}\bm{F}^{P},\bm{W}_{P}^{V}\bm{F}^{P},
\end{equation}
\begin{equation}
    \bm{F}^{P}_{attn}= \mathrm{LN}(\mathrm{Attention}(\bm{Q}_{P},\bm{K}_{P},\bm{V}_{P}) + \bm{F}^{P}),
\end{equation}
where $ \bm{W}_{P}^{Q},\bm{W}_{P}^{K},\bm{W}_{P}^{V} $ are linear projections and $ \mathrm{LN}(\cdot) $ represents layernorm layer. $ \mathrm{Attention}(\cdot) $ represents the multi-head attention, in which the results of $h$-th head can be obtained as
\begin{equation}
    \bm{F}_{attn} = \mathrm{Softmax}(\frac{\bm{Q}_{h}\bm{K}_{h}^{T}}{\sqrt{d}}),
\end{equation}
where $ d $ is the feature dimension.

Correspondingly, the image features are fed into another multi-head self-attention layer to enhance the context information as 
\begin{equation}
    \bm{Q}_{I},\bm{K}_{I},\bm{V}_{I} = \bm{W}_{I}^{Q}\bm{F}^{I},\bm{W}_{I}^{K}\bm{F}^{I},\bm{W}_{I}^{V}\bm{F}^{I},
\end{equation}
\begin{equation}
    \bm{F}^{I}_{attn}= \mathrm{LN}(\mathrm{Attention}(\bm{Q}_{I},\bm{K}_{I},\bm{V}_{I}) + \bm{F}^{I}).
\end{equation}

Then, we fuse the information of the two domains at feature level through cross-attention as
\begin{equation}
    \bm{Q}_{IP},\bm{K}_{IP},\bm{V}_{IP} = \bm{W}_{IP}^{Q}\bm{F}^{P}_{attn},\bm{W}_{IP}^{K}\bm{F}^{I}_{attn},\bm{W}_{IP}^{V}\bm{F}^{I}_{attn},
\end{equation}
\begin{equation}
    \bm{F}^{PI}_{cross}= \mathrm{LN}(\mathrm{Attention}(\bm{Q}_{IP},\bm{K}_{IP},\bm{V}_{IP}) + \bm{F}^{P}_{attn}),
\end{equation}

Note that the cross-attention is not necessary, point and image branches can work independently, which increases the flexibility of our model and allows us to train the network decoupled. 

Finally, $ \bm{F}^{PI}_{cross}$ are fed into FFN with two linear layers. 
\begin{equation}
    \bm{F}^{PI}= \mathrm{FFN}(\bm{F}^{PI}_{cross}).
\end{equation}

\begin{table*}[tp]  
\setlength{\tabcolsep}{7.0pt}
\footnotesize
\caption{Performance comparisons with state-of-the-art methods of vehicle detection on the Waymo dataset  with 202 validation sequences ($\sim$ 40k samples). $^*$: re-implemented by ourselves on OpenPCDet. }
\begin{center}
\begin{tabular}{c|l|l||cccc|cccc}
			\hline
			\multirow{2}*{Difficulty}  & \multirow{2}*{Method} & \multirow{2}*{Reference} & \multicolumn{4}{c|}{3D Detection - Vehicle} & \multicolumn{4}{c}{BEV Detection - Vehicle}\\ 
			\cline{4-11}
			&&& Overall & 0-30m & 30-50m & 50m-Inf & Overall & 0-30m & 30-50m & 50m-Inf\\
			\hline
			\multirow{10}*{LEVEL\_1} 
			& SECOND*~\cite{yan2018second} & \textit{Sensor 2018} & 72.46 & 90.30 & 70.52 & 46.93 & 89.42 & 96.58 & 88.76 & 77.55\\
			& PointPillar~\cite{lang2019pointpillars} & \textit{CVPR 2019} & 56.62 & 81.01 & 51.75 & 27.94 & 75.57 & 92.10 & 74.06 & 55.47\\
			
			& MVF~\cite{zhou2020mvf} & \textit{CoRL 2020} & 62.93 & 86.30 & 60.02 & 36.02 & 80.40 & 93.59 & 79.21 & 63.09\\
			& Pillar-OD~\cite{wang2020pillarod}& \textit{arXiv 2020} & 69.80 & 88.53 & 66.50 & 42.93 & 87.11 & 95.78 & 84.87 & 72.12\\
			& PV-RCNN~\cite{shi2020pv} & \textit{CVPR 2020} & 70.30 & 91.92 & 69.21 & 42.17 & 82.96 & 97.35 & 82.99 & 64.97\\
			& Voxel-RCNN~\cite{deng2021voxel} & \textit{AAAI 2021} & 75.59 & 92.49 & 74.09 & 53.15 & 88.19 & 97.62 & 87.34 & 77.70\\
			& LiDAR-RCNN~\cite{li2021lidar} & \textit{CVPR 2021} & 76.00 & 92.10 & 74.60 & 54.50 & 90.10 & 97.0 & 89.50 & 78.90\\
			& Pyramid R-CNN~\cite{mao2021pyramid} & \textit{ICCV 2021} & 76.30 & \textbf{92.67} & 74.91 & 54.54 & - & - & - & -\\
			& CT3D~\cite{sheng2021improving} & \textit{ICCV 2021} & 76.30 & 92.51 & 75.07 & 55.36 & 90.50 & \textbf{97.64} & 88.06 & 78.89\\
			& \textbf{FusionRCNN (Ours)} & \multicolumn{1}{c||}{-} & \textbf{78.91} & 92.38 & \textbf{77.82} & \textbf{58.81} & \textbf{91.94} & 97.12 & \textbf{91.22} & \textbf{85.22}\\
			\hline
			\multirow{6}*{LEVEL\_2} 
			& SECOND*~\cite{yan2018second}& \textit{Sensor 2018} & 64.14 & 89.04 & 64.14 & 35.98 & 82.23 & 95.63 & 83.26 & 64.29\\
			& PV-RCNN~\cite{shi2020pv} & \textit{CVPR 2020}  & 65.36 & 91.58 & 65.13 & 36.46 & 77.45 & 94.64 & 80.39 & 55.39\\
			& Voxel-RCNN~\cite{deng2021voxel} &\textit{AAAI 2021} & 66.59 & 91.74 & 67.89 & 40.80 & 81.07 & 96.99 & 81.37 & 63.26\\
			& LiDAR-RCNN~\cite{li2021lidar} & \textit{CVPR 2021} & 68.30 & 91.30 & 68.50 & 42.40 & 81.70 & 94.30 & 82.30 & 65.80\\
			& CT3D~\cite{sheng2021improving} & \textit{ICCV 2021} & 69.04 & \textbf{91.76} & 68.93 & 42.60 & 81.74 & \textbf{97.05} & 82.22 & 64.34\\
			& \textbf{FusionRCNN (Ours)} & \multicolumn{1}{c||}{-} & \textbf{70.33} & 91.22 & \textbf{71.47} & \textbf{46.21} & \textbf{84.39} & 96.22 & \textbf{86.15} & \textbf{70.18}\\
			\hline
		\end{tabular}
	\end{center}
	\label{table:waymo}
\end{table*}
\vspace{-0.2cm}

In the encoding layer, we adopt a novel fusion strategy to promote the complementary of the two modalities. The rich semantic information of image will be integrated into the point features. Correspondingly, the object structure information extracted from point branches can also guide the aggregation of image features to reduce the impact of occlusion and other situations. In our fusion encoder, we stack several encoding layers to ensure full feature fusion.

\subsection{Decoder} \label{decoding layer}
The encoded fusion features are fed into the decoding layers to obtain the features of the final box. We initialize a learnable query embedding $ \bm{E} $ with $ d $ channels as a query, in which the encoded features are used as keys and values.
\begin{equation}
    \bm{Q}_{D},\bm{K}_{D},\bm{V}_{D} = \bm{W}_{D}^{Q}\bm{E},\bm{W}_{D}^{K}\bm{F}^{PI},\bm{W}_{D}^{V}\bm{F}^{PI},
\end{equation}
\begin{equation}
    \bm{E}' = \mathrm{LN}(\mathrm{Attention}(\bm{Q}_{D},\bm{K}_{D},\bm{V}_{D}) + \bm{E}),
\end{equation}
\begin{equation}
    \bm{E}'' = \mathrm{FFN}(\bm{E}'),
\end{equation}
where $ \bm{F}^{PI} $ is the output fusion features from fusion encoding layers. The decoder module is also composed of several decoding layers.

\subsection{Objectives}
We train our model by end-to-end strategy. The overall loss is the sum of the RPN loss and the second stage network loss. RPN loss adopts the loss of the original network (SECOND~\cite{yan2018second}), and the newly introduced second stage loss includes confidence loss $ L_{conf} $ and regression loss $ L_{reg} $,
\begin{equation}
    L = L_{RPN} + L_{conf} + L_{reg}.
\end{equation}

We employ the binary cross entropy loss as the L to guide the prediction of positive samples and negative samples as
\begin{equation}
    L_{conf} = -y \log(\hat{s}) - (1-y) \log(1-\hat{s}).
\end{equation}
The division of positive and negative samples is based on IoU as
\begin{equation}
    y = \left\{
    \begin{aligned}
    1, \ \mathrm{IoU} \ge t & \\
    0, \ \mathrm{IoU} < t &
    \end{aligned}
    \right.\ \ ,
\end{equation}
where $ t $ is a threshold of IoU. For positive samples, the regression loss is composed of smooth L1 loss of all parameters of bounding box as
\begin{equation}
    L_{reg} = \sum_{p \in x,y,z,l,h,w,\theta} L_{smooth-L1}(\hat{p}, p),
\end{equation}
where $ \hat{p}, p $ represent the parameters of predictions and aligned ground truth boxes respectively.

\section{Experiments}
We evaluate FusionRCNN on both KITTI~\cite{geiger2013vision,geiger2012we} and Waymo Open Dataset~\cite{sun2020scalability}, and conduct extensive ablation studies to validate our design choices.

\subsection{Implementation Details}
\noindent\textbf{Model setup.}
We implement our network by open-sourced OpenPCDet~\cite{openpcdet2020}. We employ SECOND~\cite{yan2018second} as the RPN and follow the settings in OpenPCDet. For RoI head, we adopt ResNet50 pretrained on ImageNet\cite{krizhevsky2012imagenet} as image backbone and keep its weight frozen during training to save time, the highest resolution output of FPN is selected as the feature map. For each RoI, the expanding radio $ k $ is 2,  we sample 256 point clouds, and the corresponding projected image region is converted to 7$\times$7 resolution by RoIPooling. In addition, the number of encoding layers is set to 3 and the number of decoding layers is set to 1 to balance performance and efficiency.

\noindent\textbf{Training details.}
The network is trained end-to-end on 8 Tesla V100 GPUs. On the Waymo Open Dataset, we apply Adam optimizer and the cycle decay strategy, the learning rate is 0.0008. Following CT3D\cite{sheng2021improving}, we train the model for 80 epochs.On KITTI, we apply the same training strategy, and train 100 epochs with a learning rate of 0.003, Moreover, we design several kinds of data augmentation \emph{i.e.} flip, rotation and scaling supporting both images and point clouds.

\begin{figure*}
\begin{center}
\includegraphics[width=0.98\textwidth]{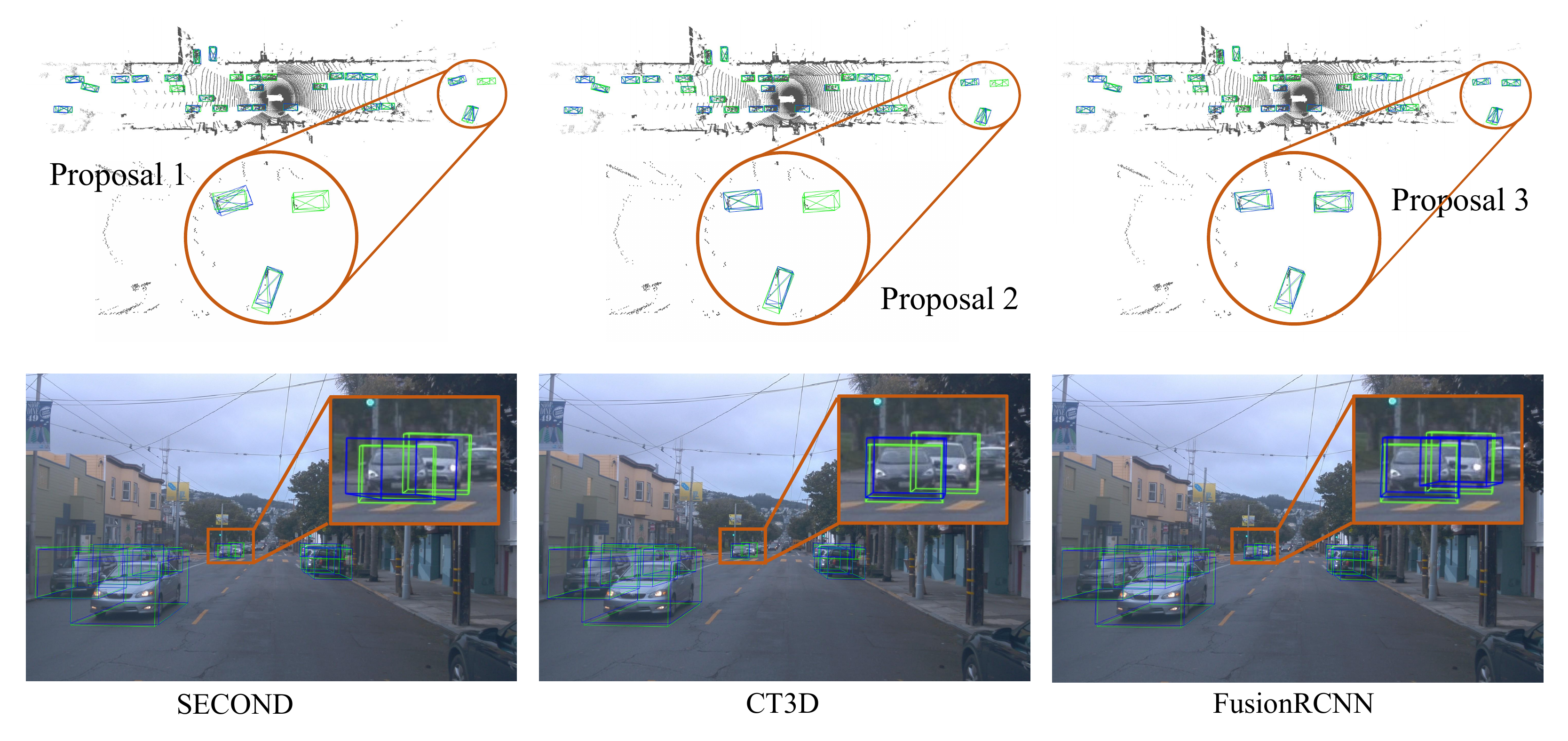}
\end{center}
\setlength{\abovecaptionskip}{1pt}  
\setlength{\belowcaptionskip}{0pt}
   \vspace{-3mm}
   \caption{Qualitative comparison between LiDAR-based two-stage detector (CT3D) and our FusionRCNN on the Waymo Open Dataset. Green boxes and Blue boxes are  ground-truth and prediction, respectively. Three proposal vehicles in red circle are zoom-in and visualize on 2D images and 3D point clouds. Our FusionRCNN works better than CT3D with only LiDAR input in long range detection. }
\label{fig:vis}
\end{figure*}

\subsection{Results on Waymo}
\noindent\textbf{Data and metrics.}
Waymo Open Dataset is a large-scale outdoor public dataset for autonomous driving research, which contains RGB images from five high-resolution cameras and 3D point clouds from five LiDAR sensors. The whole dataset consists of 798 scenes (20s fragment) for training and 202 scenes for validation and 150 for testing. 
The measures are reported based on the distances from 3D objects to sensor, \emph{i.e.}, 0-30m, 30-50m and \textgreater50m, respectively. These metrics are further divided into two difficulty levels: LEVEL1 for 3D boxes with more than 5 LiDAR points and LEVEL2 for boxes with at least 1 LiDAR point.Remarkably, the cameras in Waymo only cover around 250-degrees but not 360-degrees horizontally. Our framework can adapt to this situation. All models are trained on 20\% Waymo dataset.

\noindent\textbf{Main results.}
We first evaluate the performance of FusionRCNN on the large public Waymo Open Dataset. \cref{table:waymo} reports the results of vehicle detection with 3D and BEV AP on validation sequences. Note that with the strong SECOND~\cite{yan2018second} baseline, FusionRCNN outperforms all previous methods in both LEVEL\_1 and LEVEL\_2, leading PV-RCNN~\cite{shi2020pv} by 8.61\% mAP and Voxel-RCNN~\cite{deng2021voxel} by 3.32\% mAP on LEVEL\_1. FusionRCNN achieves 78.91\% for the commonly used LEVEL\_1 3D mAP evaluation metric, surpassing the previous state-of-the-art method CT3D~\cite{sheng2021improving} by a significant margin(2.61\% mAP). We ascribe this performance gain to our novel two-stage deep fusion design that effectively integrates geometry information from LiDAR and dense texture information from camera, which helps refine bounding box parameters and confidence scores accurately.

Additionally, we show multi-class detection results with Vehicle, Pedestrian, and Cyclist in \cref{table:waymo_multi}. After adopting FusionRCNN, we can see that the baseline model SECOND and CenterPoint~\cite{yin2021center} significantly improved small objects, \emph{i.e.}, 10.55\% mAP on Cyclist for SECOND, 6.43\% on Pedestrian for CenterPoint. \cref{table:high_iou} shows that our method surpasses other single-frame methods in the stricter eval standard(IoU threshold for 0.8), which suggests that our method works excellently in location with rich structure and texture information.

\begin{table*}  
\footnotesize
\setlength{\tabcolsep}{14.0pt}
\caption{Multi-class 3D detection results on Waymo Open validation. Both SECOND and CenterPoint baselines are implemented in OpenPCDet. "+FusionRCNN" means that we add our FusionRCNN on the baseline detector.}
\begin{center}
\begin{tabular}{c|l||cc|cc|cc}
			\hline
			\multirow{2}*{Difficulty}  & \multirow{2}*{Method}  & \multicolumn{2}{c|}{Vehicle} & \multicolumn{2}{c|}{Pedestrian}& \multicolumn{2}{c}{Cyclist}\\ 
			\cline{3-8}
			&& mAP & mAPH & mAP & mAPH & mAP & mAPH\\
			\hline
			\multirow{6}*{LEVEL\_1} 
			& SECOND~\cite{yan2018second} & 70.96 & 70.34 & 65.23 & 54.22 & 57.13 & 55.62\\
			& SECOND+FusionRCNN & 77.67 & 77.10 & 70.63 & 61.88 & 67.55 & 66.17\\
			& Improvement & \greenbf{+6.71} & \greenbf{+6.76} & \greenbf{+5.40} & \greenbf{+7.66} & \greenbf{+10.42} & \greenbf{+10.55}\\
			\cline{2-8}
			& CenterPoint~\cite{yin2021center} & 72.76 & 72.23 & 74.19 & 67.96 & 71.04 & 69.79\\
			& CenterPoint+FusionRCNN & 75.09 & 74.66 & 80.84 & 75.37 & 71.80 & 70.79\\
			& Improvement & \greenbf{+2.33} & \greenbf{+2.43} & \greenbf{+6.65} & \greenbf{+7.41} & \greenbf{+0.76} & \greenbf{+1.00}\\
			\hline
			\multirow{6}*{LEVEL\_2} 
			& SECOND & 62.58 & 62.02 & 57.22 & 47.49 & 54.97 & 53.53\\
			& SECOND+FusionRCNN & 68.84 & 68.32 & 62.67 & 54.66 & 64.67 & 63.36\\
			& Improvement & \greenbf{+6.26} & \greenbf{+6.30} & \greenbf{+5.45} & \greenbf{+7.17} & \greenbf{+9.70} & \greenbf{+9.83}\\
			\cline{2-8}
			& CenterPoint & 64.91 & 64.42 & 66.03 & 60.34 & 68.49 & 67.28\\
			& CenterPoint+FusionRCNN & 66.27 & 65.88 & 72.46 & 67.32 & 69.14 & 68.17\\
			& Improvement & \greenbf{+1.36} & \greenbf{+1.46} & \greenbf{+6.43} & \greenbf{+6.98} & \greenbf{+0.65} & \greenbf{+0.89}\\
			\hline
		\end{tabular}
	\end{center}
	\vspace{-3mm}
	\label{table:waymo_multi}
\end{table*}

\begin{table}
\small
\setlength\tabcolsep{20pt}
\caption{Results on normal and strict IoU threshold. The normal and strict thresholds for vehicles are 0.7 and 0.8, respectively, $\ast$: results from~\cite{qi2021offboard}. \vspace{-0.1cm}}
\begin{center}
\begin{tabular}{@{}l@{\ \ }c@{\ \ }c@{\ \ }c@{\ \ }c@{\ \ }c@{\ \ }}
  \toprule 
   \multirow{2}{4em}{Method} & \multirow{2}{4em}{Frames} &  \multicolumn{2}{c}{Vehicle}\\
   & & Normal & Strict \\

       \cmidrule(r){1-2}
    \cmidrule(r){3-4}

  PointPillars~\cite{lang2019pointpillars}     &    1     & 72.08 & 36.83\\  
  PV-RCNN$^\ast$~\cite{shi2020pv}         &    1     & 70.47 & 39.16\\ 
  MVF++$^\ast$~\cite{qi2021offboard}         &    1     & {74.64} & 43.30  \\ 
  SST~\cite{fan2022embracing}               &    1     & 74.22 & {44.08} \\ \hline
  \textbf{FusionRCNN (Ours)}                     &    1     & \textbf{78.91} & \textbf{47.02} \\ 
  \midrule
 \end{tabular}
\end{center}
\label{table:high_iou}
\vspace{-0.4cm}
\end{table}

\noindent\textbf{Visualization.}
Experiments on Waymo show that our method has excellent performance in long-range detection. As CT3D use the same one-stage detector as RPN, We show a qualitative comparison between FusionRCNN and CT3D which merely uses point clouds in the refinement stage. The comparison is shown in \cref{fig:vis}.

\begin{table}
\small
\setlength{\tabcolsep}{3mm}
\caption{Results on KITTI \textit{val}. Average precision with 0.7 IoU threshold and 11 recall positions are reported.\vspace{-0.2cm}}
\label{table:kitti_val}
	\footnotesize
	\begin{center}
		\begin{tabular}{l||ccc}
		\hline
			\multirow{2}*{Method}   & \multicolumn{3}{c}{3D Detection - Car}\\ 
			\cline{2-4}
			&  Easy & Mod. & Hard \\
			\hline 
			\multicolumn{4}{c}{\textit{LiDAR \& RGB}}\\
			\hline
			MV3D~\cite{chen2017mv3d} & 71.29 & 62.68 & 56.56\\
			ContFuse~\cite{liang2018cont}  & - &73.25 & -\\
			AVOD-FPN~\cite{ku2018joint}  & - & 74.44 & -\\
			F-PointNet~\cite{qi2018frustum}   &  83.76 & 70.92 &63.65\\
			PI-RCNN~\cite{xie2020pi}   &  88.27 & 78.53 & 77.75\\
			3D-CVF at SPA~\cite{yoo20203dcvf} & 89.67 & 79.88 & 78.47\\
			\hline
			\multicolumn{4}{c}{\textit{LiDAR only}}\\
			\hline
			SECOND~\cite{yan2018second} & 88.61 & 78.62 & 77.22\\
			PointPillars~\cite{lang2019pointpillars}  &  86.62 & 76.06 & 68.91\\
			STD~\cite{yang2019std} & 89.70 & 79.80 & 79.30\\
			PointRCNN~\cite{shi2019pointrcnn} & 88.88 & 78.63 & 77.38\\
			SA-SSD~\cite{he2020structure} & \textbf{90.15} & 79.91 & 78.78\\
			3DSSD~\cite{yang20203dssd} &89.71 &79.45 &78.67\\
			PV-RCNN~\cite{shi2020pv} & 89.35 & 83.69 & 78.70 \\
			Voxel-RCNN~\cite{deng2021voxel}  & 89.41 & 84.52 & 78.93\\
			Pyramid R-CNN~\cite{mao2021pyramid} & 89.37 & 84.38 & 78.84 \\
			CT3D~\cite{sheng2021improving} & 89.54 & \textbf{86.06} & 78.99 \\
			\hline
			\textbf{FusionRCNN (Ours)} & 89.90 & 85.64 & \textbf{79.32} \\
			
			\hline
		\end{tabular}
	\end{center}
\vspace{-0.5cm}
\end{table}

\begin{table}
\small
\setlength{\belowcaptionskip}{-0.3cm}
\setlength\tabcolsep{2.0pt}
\caption{Vehicle BEV detection under different distance on Waymo validation set.}
\label{tab:abla dis}
\begin{center}
\small
\begin{tabular}{l|c|ccc|c}
\hline
        Method & Overall & {0-30m} & {30-50m} & {50m-Inf} & Latency (ms)\\
        \hline
        \textbf{FusionRCNN-L} & 90.25 & 96.58 &89.24  &80.61 & 125\\
       	\textbf{FusionRCNN}  &\textbf{91.94}& \textbf{97.12} & \textbf{91.22}  & \textbf{85.22} & 185  \\
\hline

\end{tabular}
\end{center}
\vspace{-0.2cm}
\end{table}

\begin{table}
\setlength{\tabcolsep}{6.0pt}
\caption{Ablations on different one-stage detectors on Waymo validation set. \vspace{-0.2cm}}
\label{tab:diff rpn}
\begin{center}
\small
\begin{tabular}{lcc}
\hline
\multirow{2}{*}{Methods} & LEVEL\_1 & LEVEL\_2 \\
 & 3D AP / APH & 3D AP / APH \\
\hline
SECOND~\cite{yan2018second} & 72.46 / 71.87 & 64.14 / 63.60 \\
SECOND+FusionRCNN & \textbf{78.91 / 78.39} & \textbf{70.65 / 70.16}\\
\hline
PointPillar~\cite{lang2019pointpillars} & 72.27 / 71.69 & 63.85 / 63.33 \\
PointPillar+FusionRCNN  & \textbf{74.67 / 74.10} & \textbf{65.96 / 65.44} \\
\hline
CenterPoint~\cite{yin2021center} & 72.08 / 71.53 & 63.55 / 63.06 \\
CenterPoint+FusionRCNN & \textbf{77.63 / 77.16} & \textbf{69.26 / 68.83}\\
\hline
\end{tabular}
\end{center}
\vspace{-0.2cm}
\end{table}

\begin{table}
\small
\setlength\tabcolsep{16pt}
\caption{Ablation on output size of RoI image features. \vspace{-0.2cm}}
\label{tab:feature shape}
\begin{center}
\small
\begin{tabular}{ccc}
\hline
\multirow{2}{*}{Output size} & LEVEL\_1 & LEVEL\_2 \\
 & 3D AP/APH & 3D AP/APH \\
\hline
$3\times3$ & 78.88 / 78.36 & 70.63 / 70.14\\
$5\times5$ & 78.82 / 78.30 & 70.57 / 70.10 \\
$7\times7$ & \textbf{78.91} / \textbf{78.39} & \textbf{70.65} / \textbf{70.16} \\
$9\times9$ & 78.87 / 78.37 & 70.62 / 70.13 \\
\hline

\end{tabular}
\end{center}
\vspace{-0.5cm}
\end{table}

\subsection{Results on KITTI}
\noindent\textbf{Data and metrics.}
KITTI Dataset has been widely used in 3D detection tasks since its release. It contains multiple types of sensors like stereo cameras and a 64-beam Velodyne. There are 7,481 training samples commonly divided into 3,712 samples for training and 3,769 samples for validation, and 7,518 samples for testing. 
We conduct experiments on the commonly used category car whose detection IoU threshold is 0.7. We also report the results for three difficulty levels(easy, moderate and hard) according to the object size, occlusion state and truncation level.

\noindent\textbf{Main results.}
To further verify our framework, we conduct experiments on the KITTI validation set and compare with previous state-of-art methods. \cref{table:kitti_val} shows our method improves the one-stage method SECOND for all three difficulty levels with a significant margin (+1.29\% for Easy, +7.02\% for Moderate and +2.1\% for Hard) and has a great competitive with all LiDAR-based and LiDAR-Camera methods. Our FusionRCNN achieves better performance than two-stage fusion competitor PI-RCNN~\cite{xie2020pi}, which brings 7.11\% improvement on Moderate mAP. Furthermore, we compare FusionRCNN with the released method PV-RCNN~\cite{shi2020pv} and CT3D~\cite{sheng2021improving} since they share the same RPN. FusionRCNN performs better than PV-RCNN in all difficulty levels , while compared with the state-of-the-art method CT3D, our method has better performance overall, which leads CT3D by 0.36\% on Easy level and 0.33\% on Hard level with comparable result in Moderate. Remarkably, FusionRCNN achieves the AP of 79.32\%(Hard), and outperforms state-of-the-art 3D detectors. Compared with point-based two-stage methods, our novel two-stage fusion framework is better at capturing structural and contextual information effectively.

\subsection{Ablation Studies}

\noindent\textbf{Effect of LiDAR-Camera fusion.}
We investigate the effect of introducing texture information from camera images. We switch FusionRCNN to a LiDAR-based method named FusionRCNN-L by disabling the image branch in RoI Feature Extractor and cross-attention module in Fusion Encoder, then inference with the same settings. As shown in \cref{tab:abla dis}, FusionRCNN-L achieves 90.25\% mAP in Vehicle BEV detection and surpasses most of the methods in \cref{table:waymo}. By adopting LiDAR-Camera fusion, FusionRCNN gains further promotion, especially for long-range detection (50m-Inf).

\noindent\textbf{Different RPN Backbones.} we plug FusionRCNN into popular single-stage detectors, \emph{i.e.}, SECOND, PointPillar and CeterPoint to verify the generality of FusionRCNN. \cref{tab:diff rpn} shows our method improves all three baseline models with significant boosts, +6.14\%, +2.7\% and +5.55\% 3D mAP on LEVEL\_1. This benefits are from that our method utilizes a novel LiDAR-Camera fusion mechanism, leveraging structure and semantic information from LiDAR and camera images.

\noindent\textbf{RoI Feature Extractor.} Our RoI feature extractor contains a point and an image branch. 
Previous works~\cite{shi2020pv,li2021lidar,sheng2021improving} have proved that raw points have more accurate structure information to benefit local bounding box contextual information extraction. We mainly conduct an ablation study on image branch. Some parameters may affect the performance of image feature extraction and in turn detection performance. We test with different output size $S$ of RoI image features in \cref{tab:feature shape}.
We find that these settings have little impact on image extraction branch. One possible explanation is that LiDAR and image features fuse dynamically in our fusion encoding layer, and the image features contribute to category classification with high-level contextual information.

\section{Conclusion}
In this work, we propose a novel two-stage multi-modality 3D detector named FusionRCNN, which successfully integrates LiDAR point cloud and camera image information in the regions of interest. FusionRCNN leverages a well-designed  attention mechanism to achieve Set-to-Set fusion, and thus becomes more robust to the LiDAR-Camera calibration noise. We show that FusionRCNN outperforms state-of-the-art two-stage 3D detectors both on Waymo Open Dataset and KITTI dataset, which is plug-and-play and has enormous potential to boost all existing one-stage 3D detectors.  


\bibliographystyle{IEEEtran}
\bibliography{main}

\end{document}